\documentclass[runningheads]{llncs}
\usepackage{graphicx}
\usepackage[misc,geometry]{ifsym}
\usepackage{graphicx}
\usepackage{multirow}
\usepackage{algorithm}
\usepackage{algorithmic}
\usepackage{makecell}
\usepackage{tabularx}
\usepackage{multirow}
\usepackage{amsmath}
\usepackage{xcolor}
\usepackage{blkarray, bigstrut}
\usepackage{hyperref}
\hypersetup{colorlinks=true,citecolor=blue}
\usepackage{amsfonts}
\usepackage{siunitx}
\usepackage{dsfont}
\usepackage{lipsum}
\usepackage{adjustbox}
\usepackage{float}
\usepackage{booktabs}
\usepackage{cite}
\usepackage{bm}
\usepackage{pifont}
\usepackage{relsize}

\newcommand\eg {{\it e.g., }}
\newcommand\st {{\it s.t., }}

\newcommand\ie {{\it i.e., }}
\newcommand*\mystrut[1]{\vrule width0pt height0pt depth#1\relax}
\begin{document}
\title{
Distilling BlackBox to Interpretable models for Efficient Transfer Learning
}
%
%
\author{Shantanu Ghosh\inst{1}{$^{(\textrm{\Letter})}$}\and
Ke Yu\inst{2}\and
Kayhan Batmanghelich\inst{1}}

\authorrunning{S. Ghosh et al.}
%
\institute{
Department of Electrical and Computer Engineering, Boston University, Boston, MA, USA \\
\email{shawn24@bu.edu}\\
\and
Intelligent Systems Program, University of Pittsburgh, Pittsburgh, PA, USA 
}

\maketitle              
\begin{abstract}
    Building generalizable AI models is one of the primary challenges in the healthcare domain.
While radiologists rely on generalizable descriptive rules of abnormality, Neural Network (NN) models suffer even with a slight shift in input distribution (\eg scanner type).
Fine-tuning a model to transfer knowledge from one domain to another requires a significant amount of labeled data in the target domain. 
In this paper, we develop an interpretable model that can be efficiently fine-tuned to an unseen target domain with minimal computational cost.
We assume the interpretable component of NN to be approximately domain-invariant.
However, interpretable models typically underperform compared to their Blackbox (BB) variants. We start with a BB in the source domain and distill it into a \emph{mixture} of shallow interpretable models using human-understandable concepts. 
As each interpretable model covers a subset of data, a mixture of interpretable models achieves comparable performance as BB.
Further, we use the pseudo-labeling technique from semi-supervised learning (SSL) to learn the concept classifier in the target domain, followed by fine-tuning the interpretable models in the target domain.
We evaluate our model using a real-life large-scale chest-X-ray (CXR) classification dataset.
The code is available at: \url{https://github.com/batmanlab/MICCAI-2023-Route-interpret-repeat-CXRs}.
\end{abstract}

%
%
\section{Introduction}
Model generalizability is one of the main challenges of AI, especially in high stake applications such as healthcare. While NN models achieve state-of-the-art (SOTA) performance in disease classification~\cite{irvin2019chexpert, rajpurkar2017chexnet, yu2022anatomy}, they are brittle to small shifts in the data distribution~\cite{guan2021domain} caused by a change in acquisition protocol or scanner type~\cite{yan2020mri}. Fine-tuning all or some layers of a NN model on the target domain can alleviate this problem~\cite{chu2016best}, but it requires a substantial amount of labeled data and be computationally expensive~\cite{wang2017growing, kandel2020deeply}. In contrast, radiologists follow fairly generalizable and comprehensible rules. Specifically, they search for patterns of changes in anatomy to read abnormality from an image and apply logical rules for specific diagnoses. This approach is transparent and closer to an interpretable-by-design approach in AI. We develop a method to extract a mixture of interpretable models based on clinical concepts, similar to radiologists' rules, from a pre-trained NN. Such a model is more data- and computation-efficient than the original NN for fine-tuning to a new distribution.

 Standard interpretable by design method~\cite{rudin2022interpretable} finds an interpretable function (\eg linear regression or rule-based) between human-interpretable concepts and final output~\cite{koh2020concept}. A concept classifier~\cite{sarkar2021inducing, zarlenga2022concept} detects the presence or absence of concepts in an image. In medical images, previous research uses TCAV scores~\cite{kim2017interpretability} to quantify the role of a concept on the final prediction~\cite{yeche2019ubs, graziani2020concept, clough2019global}, but the concept-based interpretable models have been mostly unexplored.
Recently Posthoc Concept Bottleneck models (PCBMs)~\cite{yuksekgonul2022post} identify concepts from the embeddings of BB. However, the common design choice amongst those methods relies on a single interpretable classifier to explain the entire dataset, cannot capture the diverse sample-specific explanations, and performs poorly than their BB variants.

\textbf{Our contributions.}
This paper proposes a novel data-efficient interpretable method that can be transferred to an unseen domain. Our interpretable model is built upon human-interpretable concepts and can provide sample-specific explanations for diverse disease subtypes and pathological patterns. Beginning with a BB in the source domain, we progressively extract a mixture of interpretable models from BB. Our method includes a set of selectors routing the explainable samples through the interpretable models. The interpretable models provide First-order-logic (FOL) explanations for the samples they cover. The remaining unexplained samples are routed through the residuals until they are covered by a successive interpretable model. We repeat the process until we cover a desired fraction of data. Due to class imbalance in large CXR datasets, early interpretable models tend to cover all samples with disease present while ignoring disease subgroups and pathological heterogeneity. We address this problem by estimating the class-stratified coverage from the total data coverage. We then finetune the interpretable models in the target domain. The target domain lacks concept-level annotation since they are expensive. Hence, we learn a concept detector in the target domain with a pseudo labeling approach~\cite{lee2013pseudo} and finetune the interpretable models. Our work is the first to apply concept-based methods to CXRs and transfer them between domains. 


\label{sec:intro}


\section{Methodology}
\begin{figure*}[t]
\begin{center}
\includegraphics[width=\linewidth]{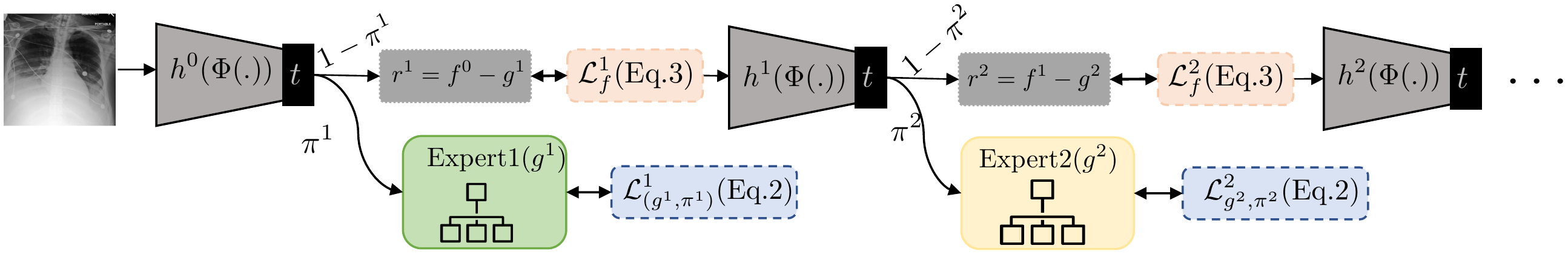}
\caption{Schematic view of our method. Note that $f^k(.) = h^k(\Phi(.))$. At iteration $k$, the selector \emph{routes} each sample either towards the expert $g^k$ with probability $\pi^k(.)$ or the residual $r^k = f^{k-1} - g^k$ with probability $1-\pi^k(.)$. $g^k$ generates FOL-based explanations for the samples it covers. Note $\Phi$ is fixed across iterations.}
\label{fig:schematic}
\end{center}
\end{figure*}

\noindent\textbf{Notation.}
Assume $f^0: \mathcal{X} \rightarrow \mathcal{Y}$ is a BB, trained on a dataset $\mathcal{X} \times\mathcal{Y} \times \mathcal{C}$, with $\mathcal{X}$, $\mathcal{Y}$, and $\mathcal{C}$ being the images, classes, and concepts, respectively; $f^0=h^0 \circ \Phi$, where $\Phi$ and  $h^0$ is the feature extractor and the classifier respectively. Also, $m$ is the number of class labels. This paper focuses on binary classification (having or not having a disease), so $m=2$ and $\mathcal{Y} \in \{0, 1\}$. Yet, it can be extended to multiclass problems easily. Given a learnable projection~\cite{pmlr-v202-ghosh23c, ghosh2023tackling}, $t: \Phi \rightarrow \mathcal{C}$, our method learns three functions: (1) a set of selectors ($\pi: \mathcal{C}\rightarrow \{0, 1\}$) routing samples to an interpretable model or residual, (2) a set of interpretable models ($g: \mathcal{C} \rightarrow \mathcal{Y}$), and (3) the residuals. The interpretable models are called ``experts" since they specialize in a distinct subset of data defined by that iteration's coverage $\tau$ as shown in SelectiveNet~\cite{rabanser2022selective}. Fig.~\ref{fig:schematic} illustrates our method.

\subsection{Distilling BB to the mixture of interpretable models}
\noindent\textbf{Handling class imbalance.} For an iteration $k$, we first split the given coverage $\tau^k$ to stratified coverages per class as $\{\tau^k_m = w_m \cdot \tau^k; w_m=N_m/N, \forall m\}$, where $w_m$ denotes the fraction of samples belonging to the $m^{th}$ class; $N_m$ and $N$ are the samples of $m^{th}$ class and total samples, respectively. 

\noindent \textbf{Learning the selectors.}
At iteration $k$, the selector $\pi^k$ \emph{routes} $i^{th}$ sample to the expert ($g^k$) or residual ($r^k$) with probability $\displaystyle \pi^k(\boldsymbol{c_i})$ and $\displaystyle 1 - \pi^k(\boldsymbol{c_i})$ respectively. 
For coverages $\{\tau^k_m, \forall m\}$, we learn $g^k$ and $\pi^k$ jointly by solving the 
loss:
\begin{align}
\label{equ: optimization_g}
\theta_{s^k}^*, \theta_{g^k}^* = & \operatorname*{arg\,min}_{\theta_{s^k}, \theta_{g^k}} \mathcal{R}^k\Big(\pi^k(.; \theta_{s^k}), \displaystyle g^k(.; \theta_{g^k}) \Big) 
~~\text{s.t.}~\zeta_m\big(\pi^k(.; \theta_{s^k})\big) \geq \tau^k_m ~~\forall m,
\end{align}
where $\theta_{s^k}^*, \theta_{g^k}^*$ are the optimal parameters for $\pi^k$ and $g^k$, respectively.
$\mathcal{R}^k$ is the overall selective risk, defined as,
$\mathcal{R}^k(\displaystyle \pi^k, \displaystyle g^k) = \mathlarger{\sum}_m\frac{\frac{1}{N_m}\sum_{i=1}^{N_m}\mathcal{L}_{(g^k, \pi^k)}^k\big(\boldsymbol{x_i}, \boldsymbol{c_i}\big)}{\zeta_m(\pi^k)}$ ,
where $\zeta_m(\pi^k) = \frac{1}{N_m}\sum_{i=1}^{N_m}\pi^k(\boldsymbol{c_i})$ is the empirical mean of samples of $m^{th}$ class selected by the selector for the associated expert $\displaystyle g^k$. We define $\mathcal{L}_{(g^k, \pi^k)}^k$ in the next section.
The selectors are neural networks with sigmoid activation. At inference time, $\pi^k$ routes a sample  to $\displaystyle g^k$ if and only if $\pi^k(.)\geq 0.5$. 

\noindent \textbf{Learning the experts.}
\label{learn_explert}
For iteration $k$, the loss $\mathcal{L}_{(g^k, \pi^k)}^k$ distills the expert $g^k$ from $f^{k-1}$, BB of the previous iteration by solving the following loss:
\begin{equation}
\label{equ: g_k}
\resizebox{0.8\textwidth}{!}{$
\mathcal{L}_{(g^k, \pi^k)}^k\big(\boldsymbol{x_i}, \boldsymbol{c_i}\big) = \underbrace{\mystrut{2.6ex}\ell\Big(f^{k - 1}(\boldsymbol{x_i}), g^k(\boldsymbol{c_i})\Big)\pi^k(c_i) }_{\substack{\text{trainable component} \\ \text{for current iteration $k$}}}\underbrace{\prod_{j=1} ^{k - 1}\big(1 - \pi^j(\boldsymbol{c_i})\big)}_{\substack{\text{fixed component trained} \\ \text{in the previous iterations}}},
$}
\end{equation}
where $\pi^k(\boldsymbol{c_i})\prod_{j=1} ^{k - 1}\big(1 - \pi^j(\boldsymbol{c_i}) \big)$ is the cumulative probability of the sample covered by the residuals for all the previous iterations from $1, \cdots, k-1$ (\ie $\prod_{j=1} ^{k - 1}\big(1 - \pi^j(\boldsymbol{c_i}) \big)$\big) and the expert $g^k$ at iteration $k$ \big(\ie $\pi^k(\boldsymbol{c_i})$\big). 

\noindent \textbf{Learning the Residuals.}
After learning $g^k$, we calculate the residual as, $r^k (x_i, c_i) = f^{k-1}(x_i) - g^k(c_i)$ (difference of logits). We fix $\Phi$ and optimize the following loss to update $h^k$ to specialize on those samples not covered by $g^k$, effectively creating a new BB $f^k$ for the next iteration $(k+1)$:
\begin{equation}
\label{equ: residual}
\mathcal{L}_f^k(\boldsymbol{x_j}, \boldsymbol{c_j}) = \underbrace{\mystrut{2.6ex}\ell\big(r^k(\boldsymbol{x_j}, \boldsymbol{c_j}), f^k(\boldsymbol{x_j})\big)}_{\substack{\text{trainable component} \\ \text{for iteration $k$}}} \underbrace{\mystrut{2.6ex}\prod_{i=1} ^{k}\big(1 - \pi^i(\boldsymbol{c_j})\big)}_{\substack{\text{non-trainable component} \\ \text{for iteration $k$}}}
\end{equation}
 We refer to all the experts as the Mixture of Interpretable Experts (MoIE-CXR). We denote the models, including the final residual, as MoIE-CXR+R. Each expert in MoIE-CXR constructs sample-specific FOLs using the optimization strategy and algorithm discussed in~\cite{pmlr-v202-ghosh23c}.

\subsection{Finetuning to an unseen domain} 
We assume the MoIE-CXR-identified concepts to be generalizable to an unseen domain. So, we learn the projection $t_t$ for the target domain and compute the pseudo concepts using SSL~\cite{lee2013pseudo}. Next, we transfer the selectors, experts, and final residual ($\{\pi^k_s, g^k_s\}_{k=1}^K$ and $f^K_s$) from the source to a target domain with limited labeled data and computational cost.
Algorithm~\ref{algo: domain_transfer} details the procedure.

\begin{algorithm}[H]
   \caption{Finetuning to an unseen domain.}
   \label{algo: domain_transfer}
\begin{algorithmic}[1]
   \STATE {\bfseries Input:} Learned selectors, experts, and final residual from source domain: $\{\pi^k_s, g^k_s\}_{k=1}^K$ and $f^K_s$ respectively, with $K$ as the number of experts to transfer. BB of the source domain: $f_s^0=h^0_s(\Phi_s)$. Source data: $\mathcal{D}_s = \{\mathcal{X}_s, \mathcal{C}_s, \mathcal{Y}_s\}$. Target data: $\mathcal{D}_t = \{\mathcal{X}_t, \mathcal{Y}_t\}$. Target coverages $\{\tau_k\}_{k=1}^K$.
   \STATE {\bfseries Output:} Experts $\{\pi^k_t, g^k_t\}_{k=1}^K$ and final residual $f^K_t$ of the target domain.
   \STATE Randomly select $n_t\ll N_t$ samples out of $N_t=|\mathcal{D}_t|$.
   \STATE Compute the pseudo concepts for the correctly classified samples in the target domain using $f^0_s$, as, $\boldsymbol{c_t^i} = t_s\big(\Phi_s(\boldsymbol{x_s}^i)\big)$ \st $y_t^i=f^0_s(\boldsymbol{x_t}^i)$, $i=1 \cdots~n_t$
   \STATE \label{step:psudo-label} Learn the projection function $t_t$ for target domain semi-supervisedly~\cite{lee2013pseudo} using the pseudo labeled samples $\{\boldsymbol{x}_t^i, \boldsymbol{c}_t^i\}_{i=1}^{n_t}$ and unlabeled samples $\{\boldsymbol{x}_t^i\}_{i=1}^{N_t-n_t}$.
   \STATE Complete the triplet for the target domain \{$\mathcal{X}_t, \mathcal{C}_t, \mathcal{Y}_t$\}, where $\boldsymbol{c}_t^i=t_t(\Phi_s(\boldsymbol{x}_t^i))$, $i=1\cdots~N_t$.
   \STATE Finetune $\{\pi^k_s, g^k_s\}_{k=1}^K$ and $f^K_s$ to obtain $\{\pi^k_t, g^k_t\}_{k=1}^K$ and $ f^K_t$ using equations~\ref{equ: optimization_g},~\ref{equ: g_k} and~\ref{equ: residual} respectively for 5 epochs. $\{\pi^k_t, g^k_t\}_{k=1}^K$ and $\big\{ \{\pi^k_t, g^k_t\}_{k=1}^K, f_t^K \big\}$ represents MoIE-CXR and MoIE-CXR + R for the target domain.
\end{algorithmic}
\end{algorithm}
\label{sec:method}

\section{Experiments}
We perform experiments to show that MoIE-CXR 1) captures a diverse set of concepts, 2) does not compromise BB's performance, 3) covers ``harder'' instances with the residuals in later iterations resulting in their drop in performance, 4) is finetuned well to an unseen domain with minimal computation. 

\begin{figure*}[ht]
\begin{center}
\includegraphics[width=\linewidth]{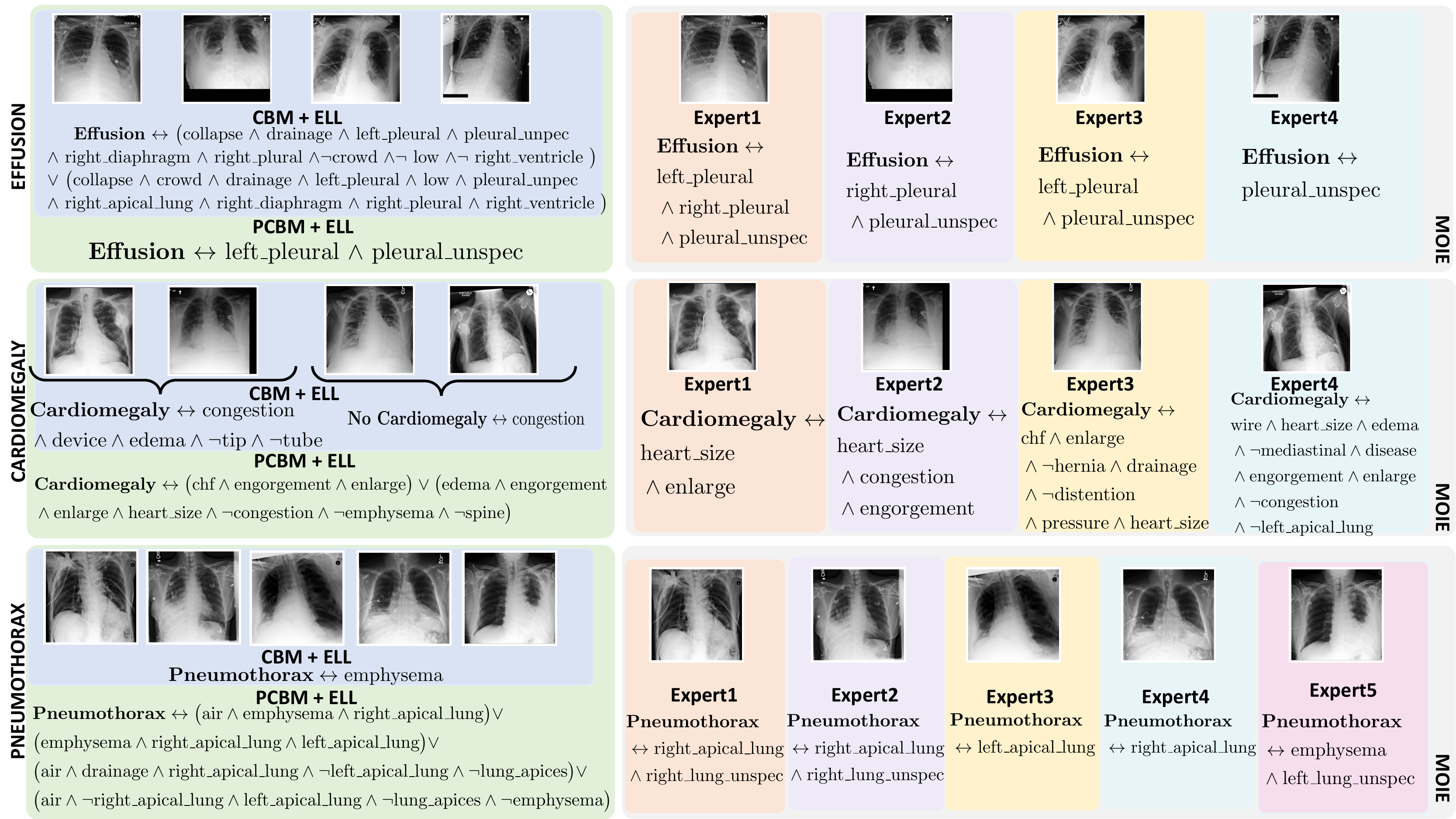}
\caption{Qualitative comparison of MoIE-CXR discovered concepts with the baselines.}
\label{fig:qual}
\end{center}
\end{figure*}

\noindent \textbf{Experimental Details.} We evaluate our method using  220,763 frontal images from the MIMIC-CXR dataset \cite{12_johnsonmimic}. We use Densenet121 \cite{huang2017densely} as BB ($f^0$) to classify cardiomegaly, effusion, edema, pneumonia, and pneumothorax, considering each to be a separate binary classification problem. We obtain 107 anatomical and observation concepts from the RadGraph’s inference dataset~\cite{10_jain2021radgraph}, automatically generated by DYGIE++~\cite{23_wadden-etal-2019-entity}. We train BB following~\cite{yu2022anatomy}. To retrieve the concepts, we utilize until the $4^{th}$ Densenet block as feature extractor $\Phi$ and flatten the features to learn $t$. We use an 80\%-10\%-10\% train-validation-test split with no patient shared across splits. We use 4, 4, 5, 5, and 5 experts for cardiomegaly, pneumonia, effusion, pneumothorax, and edema. We employ ELL~\cite{barbiero2022entropy} as $g$. Further, we only include concepts as
input to $g$ if their validation auroc exceeds 0.7. Refer to Tab. 1 in the supplementary material for the hyperparameters. 
We stop until all the experts cover at least 90\% of the data cumulatively. \noindent \textbf{Baseline.} We compare our method with 1) end-to-end CEM~\cite{zarlenga2022concept}, 2) sequential CBM~\cite{koh2020concept}, and 3) PCBM~\cite{yuksekgonul2022post} baselines, comprising of two parts: a) concept predictor $\Phi: \mathcal{X} \rightarrow \mathcal{C}$, predicting concepts from images, with all the convolution blocks; and b) label predictor, $g: \mathcal{C} \rightarrow \mathcal{Y}$, predicting labels from the concepts. We create CBM + ELL and PCBM + ELL by replacing the standard classifier with the identical $g$ of MOIE-CXR to generate FOLs~\cite{barbiero2022entropy} for the baseline.
\label{sec:experiment}

\noindent \textbf{MoIE-CXR captures diverse explanations.}
Fig.~\ref{fig:qual} illustrates the FOL explanations. Recall that the experts ($g$) in MoIE-CXR and the baselines are ELLs~\cite{barbiero2022entropy}, attributing attention weights to each concept. A concept with high attention weight indicates its high predictive significance. With a single $g$, the baselines rank the concepts in accordance with the identical order of attention weights for all the samples in a class, yielding  a generic FOL for that class. In Fig.~\ref{fig:qual}, the baseline PCBM + ELL uses \emph{left\_pleural} and \emph{pleural\_unspec} to identify effusion for all four samples. MoIE-CXR deploys multiple experts, learning to specialize in distinct subsets of a class. So different interpretable models in MoIE assign different attention weights to capture instance-specific concepts unique to each subset. In Fig.~\ref{fig:qual} expert2 relies on \emph{right\_pleural} and \emph{pleural\_unspec}, but expert4 relies only on \emph{pleural\_unspec} to classify effusion. 
The results show that the learned experts can provide more precise explanations at the subject level using the concepts, increasing confidence and trust in clinical use. 
\label{sec:qual_analysis}

\begin{table*}[h]
\caption{MoIE-CXR does not compromize the performance of BB. We 
provide the mean and standard errors of AUROC over five random seeds. For MoIE-CXR, we also report the percentage of test set samples covered by all experts as ``\emph{Coverage}''. We boldfaced our results and BB.}
\begin{adjustbox}{max width=1\textwidth, center}
\begin{tabular}{lccccccccc}
\toprule
\textbf{Model} &
  Effusion &
  Cardiomegaly &
  Edema &
  Pneumonia &
  Pneumothorax
  \\
\midrule 
  \scriptsize{Blackbox (BB)} & 
  \scriptsize{$\boldsymbol{0.92}$} & 
  \scriptsize{$\boldsymbol{0.84}$} & 
  \scriptsize{$\boldsymbol{0.89}$} & 
  \scriptsize{$\boldsymbol{0.79}$} 
  & \scriptsize{$\boldsymbol{0.91}$}
  \\
\midrule
\scriptsize{\textbf{INTERPRETABLE BY DESIGN}} \\
\scriptsize{CEM}~\cite{zarlenga2022concept} &
\scriptsize{$0.83_{\pm 1\mathrm{e}{-4}}$} &
\scriptsize{$0.75_{\pm 1\mathrm{e}{-4}}$} &
\scriptsize{$0.77_{\pm 2\mathrm{e}{-4}}$} &
\scriptsize{$0.62_{\pm 4\mathrm{e}{-4}}$} &
\scriptsize{$0.76_{\pm 3\mathrm{e}{-4}}$} &
\\
\scriptsize{CBM (Sequential)}~\cite{koh2020concept}  &  
\scriptsize{$0.78_{\pm 1\mathrm{e}{-4}}$} &
\scriptsize{$0.72_{\pm 1\mathrm{e}{-4}}$} &
\scriptsize{$0.77_{\pm 5\mathrm{e}{-4}}$} &
\scriptsize{$0.60_{\pm 1\mathrm{e}{-3}}$} &
\scriptsize{$0.75_{\pm 6\mathrm{e}{-4}}$} &
\\
\scriptsize{CBM + ELL}~\cite{koh2020concept, barbiero2022entropy}    &
\scriptsize{$0.81_{\pm 1\mathrm{e}{-4}}$} &
\scriptsize{$0.72_{\pm 1\mathrm{e}{-4}}$} &
\scriptsize{$0.79_{\pm 5\mathrm{e}{-4}}$} &
\scriptsize{$0.62_{\pm 8\mathrm{e}{-4}}$} &
\scriptsize{$0.75_{\pm 6\mathrm{e}{-4}}$} &
\\

\midrule
\scriptsize{\textbf{POSTHOC}} \\
\scriptsize{PCBM}~\cite{yuksekgonul2022post} & 
\scriptsize{$0.88_{\pm 1\mathrm{e}{-4}}$} & 
\scriptsize{$0.81_{\pm 1\mathrm{e}{-4}}$} & 
\scriptsize{$0.82_{\pm 1\mathrm{e}{-4}}$} &
\scriptsize{$0.72_{\pm 1\mathrm{e}{-4}}$} &
\scriptsize{$0.85_{\pm 7\mathrm{e}{-4}}$} &
 \\ 
\scriptsize{PCBM-h}~\cite{yuksekgonul2022post} &
\scriptsize{$0.90_{\pm 1\mathrm{e}{-4}}$} &
\scriptsize{$0.83_{\pm 1\mathrm{e}{-4}}$} & 
\scriptsize{$0.85_{\pm 1\mathrm{e}{-4}}$} &
\scriptsize{$0.77_{\pm 1\mathrm{e}{-4}}$} &
\scriptsize{$0.89_{\pm 7\mathrm{e}{-4}}$} &
\\ 
\scriptsize{PCBM + ELL}~\cite{yuksekgonul2022post, barbiero2022entropy} &
\scriptsize{$0.90_{\pm 1\mathrm{e}{-4}}$} &
\scriptsize{$0.82_{\pm 1\mathrm{e}{-4}}$} &
\scriptsize{$0.85_{\pm 1\mathrm{e}{-4}}$} &
\scriptsize{$0.75_{\pm 1\mathrm{e}{-4}}$} &
\scriptsize{$0.85_{\pm 6\mathrm{e}{-4}}$} &
\\
\scriptsize{PCBM-h + ELL}~\cite{yuksekgonul2022post, barbiero2022entropy} &
\scriptsize{$0.91_{\pm 1\mathrm{e}{-4}}$} &
\scriptsize{$0.83_{\pm 1\mathrm{e}{-4}}$} & 
\scriptsize{$0.87_{\pm 1\mathrm{e}{-4}}$} &
\scriptsize{$0.77_{\pm 1\mathrm{e}{-4}}$} &
\scriptsize{$0.90_{\pm 1\mathrm{e}{-4}}$} &
\\
 
\midrule
\scriptsize{\textbf{OURS}} \\
\scriptsize{MoIE-CXR $^{\text{(Coverage)}}$} & 
\scriptsize{$\boldsymbol{0.93^\textbf{\emph{(0.90)}}_{\pm 1\mathrm{e}{-4}}}$} &
\scriptsize{$\boldsymbol{0.85^\textbf{\emph{(0.96)}}_{\pm 1\mathrm{e}{-4}}}$} &
\scriptsize{$\boldsymbol{0.91^\textbf{\emph{(0.92)}}_{\pm 1\mathrm{e}{-4}}}$} &
\scriptsize{$\boldsymbol{0.80^\textbf{\emph{(0.97)}}_{\pm 1\mathrm{e}{-4}}}$} &
\scriptsize{$\boldsymbol{0.91^\textbf{\emph{(0.93)}}_{\pm 2\mathrm{e}{-4}}}$} &
\\ 

\scriptsize{MoIE-CXR+R} & 
\scriptsize{$\boldsymbol{0.91_{\pm 1\mathrm{e}{-4}}}$} &
\scriptsize{$\boldsymbol{0.82_{\pm 1\mathrm{e}{-4}}}$} &
\scriptsize{$\boldsymbol{0.88_{\pm 1\mathrm{e}{-4}}}$} &
\scriptsize{$\boldsymbol{0.78_{\pm 1\mathrm{e}{-4}}}$} &
\scriptsize{$\boldsymbol{0.90_{\pm 2\mathrm{e}{-4}}}$}
\\ 
\bottomrule
\end{tabular}
\end{adjustbox}
\label{tab:performance}
\end{table*}

\noindent \textbf{MoIE-CXR does not compromise BB's performance.}
\noindent \textbf{Analysing MoIE-CXR:} Tab.~\ref{tab:performance} shows that MoIE-CXR outperforms other models, including BB. Recall that MoIE-CXR refers to the mixture of all interpretable experts, excluding any residuals. As MoIE-CXR specializes in various subsets of data, it effectively discovers sample-specific classifying concepts and achieves superior performance.
In general, MoIE-CXR exceeds the interpretable-by-design baselines (CEM, CBM, and CBM + ELL) by a fair margin (on average, at least $\sim 10\% \uparrow$), especially for pneumonia and pneumothorax where the number of samples with the disease is significantly less ($\sim 750/24000$ in the testset).
\noindent\textbf{Analysing MoIE-CXR+R:}
 To compare the performance on the entire dataset, we additionally report MoIE-CXR+R, the mixture of interpretable experts with the final residual in Tab.\ref{tab:performance}. MoIE-CXR+R outperforms the interpretable-by-design models and yields comparable performance as BB. The residualized PCBM baseline, \ie PCBM-h, performs similarly to MoIE-CXR+R.
 PCBM-h rectifies the interpretable PCBM's mistakes by learning the residual with the complete dataset to resemble BB's performance. However, the experts and the final residual approximate the interpretable and uninterpretable fractions of BB, respectively. In each iteration, the residual focuses on the samples not covered by the respective expert to create BB for the next iteration and likewise. As a result, the final residual in MoIE-CXR+R covers the "hardest" examples, reducing its overall performance relative to MoIE-CXR. 
\label{sec:quant_analysis}

\begin{figure*}[t]
\begin{center}
\centerline{\includegraphics[width=\linewidth]{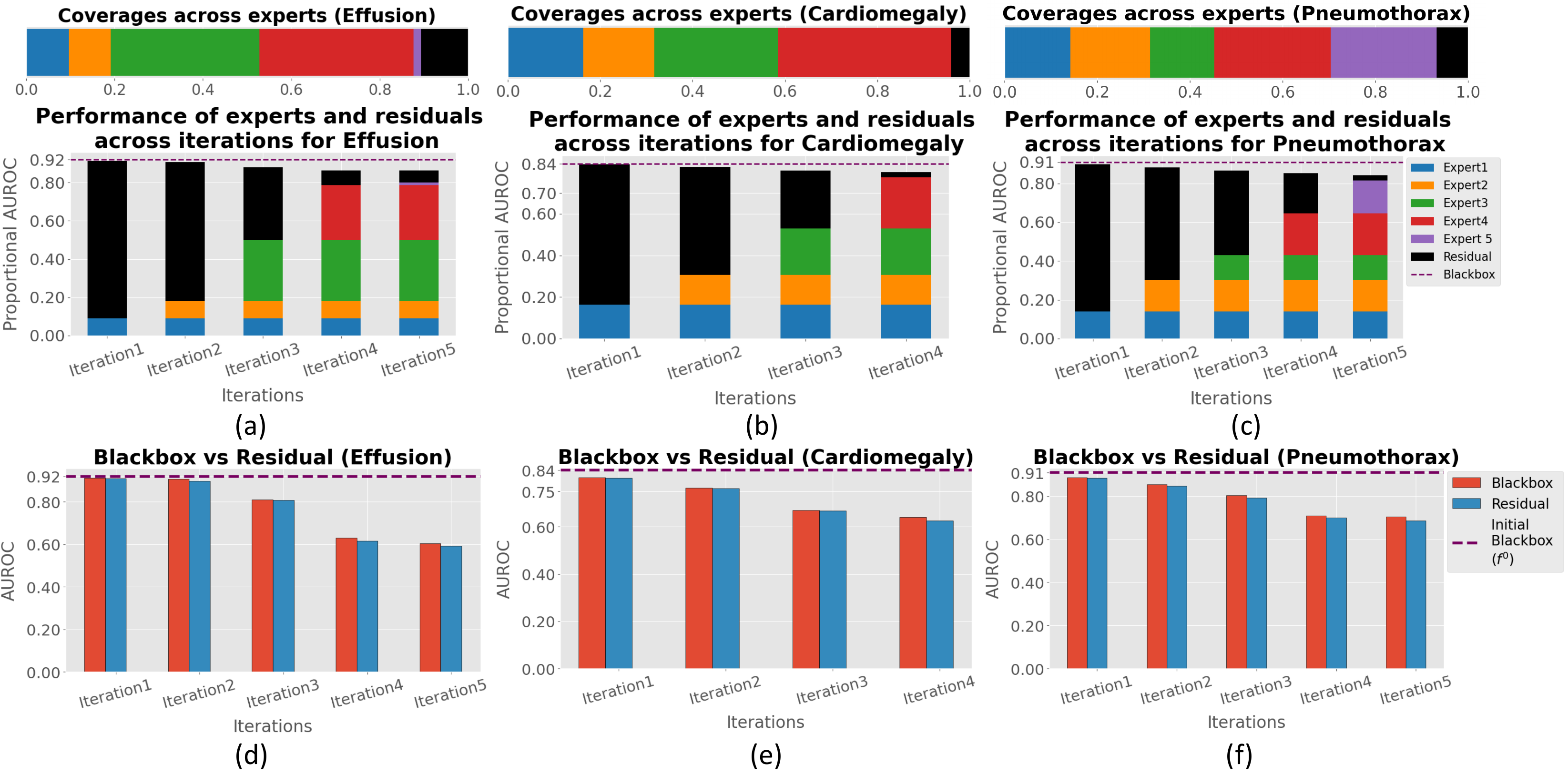}}
\caption{Performance of experts and residuals across iterations. 
\textbf{(a-c):} Coverage and proportional AUROC of the experts and residuals.
\textbf{(d-f):} Routing the samples covered by MoIE-CXR to the initial $f^0$, we compare the performance of the residuals with $f^0$.
}
\label{fig:expert_performance_cv_vit}
\end{center}
\end{figure*}

\noindent \textbf{Identification of harder samples by successive residuals.}
Fig.~\ref{fig:expert_performance_cv_vit} (a-c) reports the proportional AUROC of the experts and the residuals per iteration. The proportional AUROC is the AUROC of that model times the empirical coverage, $\zeta^k$, the mean of the samples routed to the model by the respective selector ($\pi^k$).
According to Fig.~\ref{fig:expert_performance_cv_vit}a in iteration 1, the residual (black bar) contributes more to the proportional AUROC than the expert1 (blue bar) for effusion with both achieving a cumulative proportional AUROC~$\sim$ 0.92. All the final experts collectively extract the entire interpretable component from BB $f^0$ in the final iteration, resulting in their more significant contribution to the cumulative performance. In subsequent iterations, the proportional AUROC decreases as the experts are distilled from the BB of the previous iteration. The BB is derived from the residual that performs progressively worse with each iteration. The residual of the final iteration covers the ``hardest'' samples. Tracing these samples back to the original BB $f^0$, $f^0$ underperforms on these samples (Fig.~\ref{fig:expert_performance_cv_vit} (d-f)) as the residual.

\label{sec:residual_performance}

\begin{figure*}[t]
\vskip 0.2in
\begin{center}
\centerline{\includegraphics[width=\linewidth]{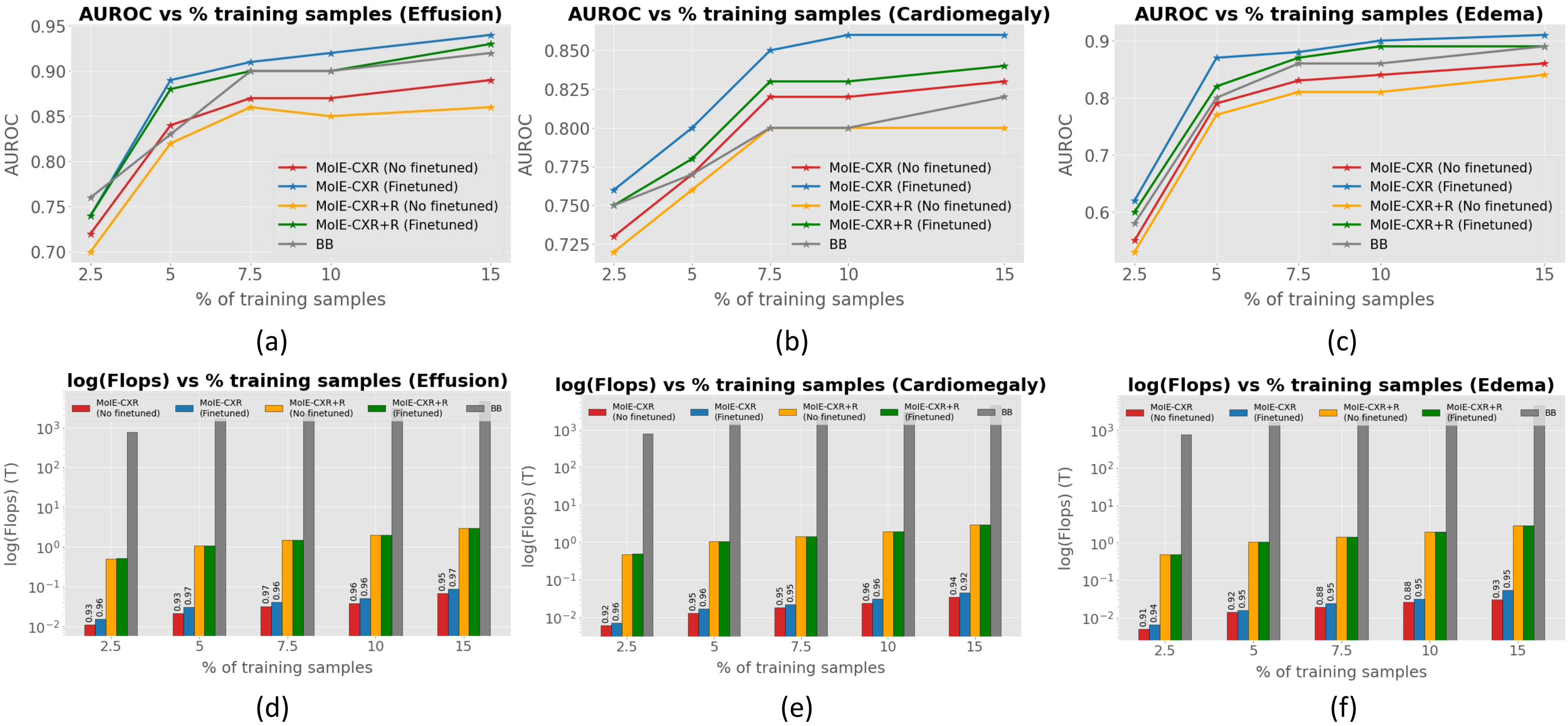}}
\caption{Transfering the first 3 experts of MoIE-CXR trained on MIMIC-CXR to Stanford-CXR. With varying \% of training samples of Stanford CXR, \textbf{(a-c):}  reports AUROC of the test sets, \textbf{(d-g)} reports computation costs in terms of $\log\text{(Flops) (T)}$.
We report the coverages in Stanford-CXR on top of the ``finetuned'' and ``No finetuned'' variants of MoIE-CXR (red and blue bars) in \textbf{(d-g)}.
}
\label{fig:domain_generelization}
\end{center}
\end{figure*}

\noindent \textbf{Applying MoIE-CXR to the unseen domain.} In this experiment, we utilize Algo.~\ref{algo: domain_transfer} to transfer MoIE-CXR trained on MIMIC-CXR dataset to Stanford Chexpert~\cite{irvin2019chexpert} dataset for the diseases -- effusion, cardiomegaly and edema. Using 2.5\%, 5\%, 7.5\%, 10\%, and 15 \% of training data from the Stanford Chexpert dataset, we employ two variants of MoIE-CXR where we (1) train only the selectors ($\pi$) without finetuning the experts ($g$) (``No finetuned'' variant of MoIE-CXR in Fig.~\ref{fig:domain_generelization}), and (2) finetune $\pi$ and $g$ jointly for only 5 epochs (``Finetuned'' variant of MoIE-CXR and MoIE-CXR + R in Fig.~\ref{fig:domain_generelization}). Finetuning $\pi$ is essential to route the samples of the target domain to the appropriate expert. As later experts cover the ``harder'' samples of MIMIC-CXR, we only transfer the experts of the first three iterations (refer to Fig.~\ref{fig:expert_performance_cv_vit}). To ensure a fair comparison, we finetune (both the feature extractor $\Phi$ and classifier $h^0$) BB: $f^0 = h^0 \circ \Phi$ of MIMIC-CXR with the same training data of Stanford Chexpert for 5 epochs. Throughout this experiment, we fix $\Phi$ while finetuning the final residual in MoIE+R as stated in Eq.~\ref{equ: residual}. Fig.~\ref{fig:domain_generelization} displays the performances of different models and the computation costs in terms of Flops. The Flops are calculated as, Flop of (forward propagation + backward propagation) $\times$ (total no. of batches) $\times$ (no of training epochs). The finetuned MoIE-CXR outperforms the finetuned BB (on average $\sim 5\% \uparrow$ for effusion and cardiomegaly).
As experts are simple models~\cite{barbiero2022entropy} and accept only low dimensional concept vectors compared to BB, the computational cost to train MoIE-CXR is significantly lower than that of BB (Fig.~\ref{fig:domain_generelization} (d-f)). Specifically, BB requires $\sim$ 776T flops to be finetuned on 2.5\% of the training data of Stanford CheXpert, whereas MoIE-CXR requires $\sim$ 0.0065T flops. As MoIE-CXR discovers the sample-specific domain-invariant concepts, it achieves such high performance with low computational cost than BB. 

\label{sec:domain_genralization}

\section{Conclusion}
This paper proposes a novel iterative interpretable method that identifies instance-specific concepts without losing the performance of the BB and is effectively fine-tuned in an unseen target domain with no concept annotation, limited labeled data, and minimal computation cost. Also, as in the prior work, MoIE-captured concepts may not showcase a causal effect that can be explored in the future.

\label{sec:conclusion}

\section{Acknowledgement}
This work was partially supported by NIH Award Number 1R01HL141813-01 and the Pennsylvania Department of Health. We are grateful for the computational resources provided by Pittsburgh Super Computing grant number TG-ASC170024.

\label{sec:ack}


%
%
%
\bibliographystyle{splncs04}
\bibliography{main}

\setcounter{table}{0}
\setcounter{figure}{0}
\newpage
\section*{Supplementary materials}
\begin{figure*}[h]
\begin{center}
\centerline{\includegraphics[width=\linewidth]{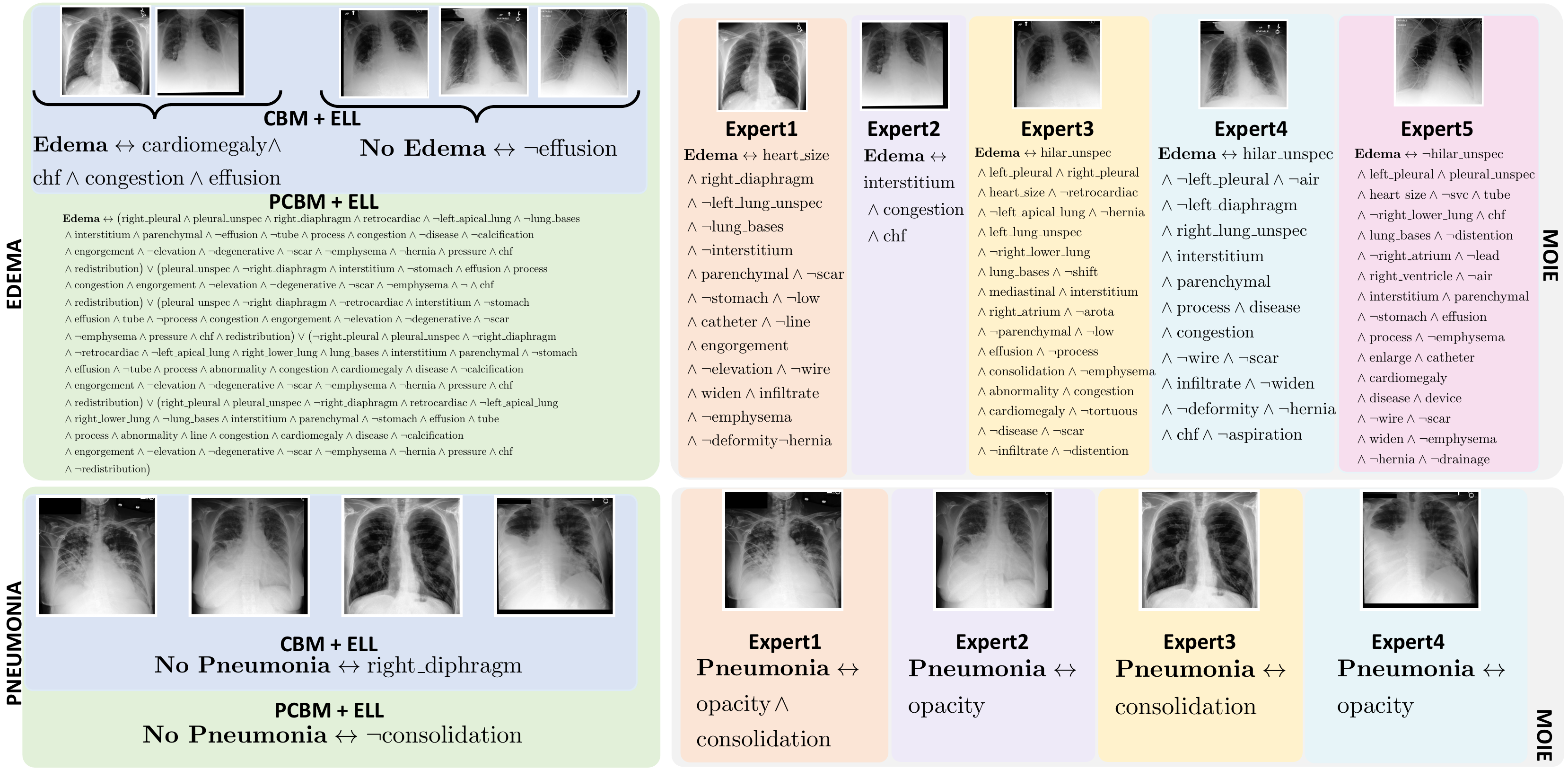}}
\caption{Qualitative comparison of MoIE-CXR discovered concepts with the baseline for edema and pneumonia.}
\label{fig:expert_performance_cv_vit}
\end{center}
\end{figure*}

\begin{figure*}[h]
\begin{center}
\centerline{\includegraphics[width=\linewidth]{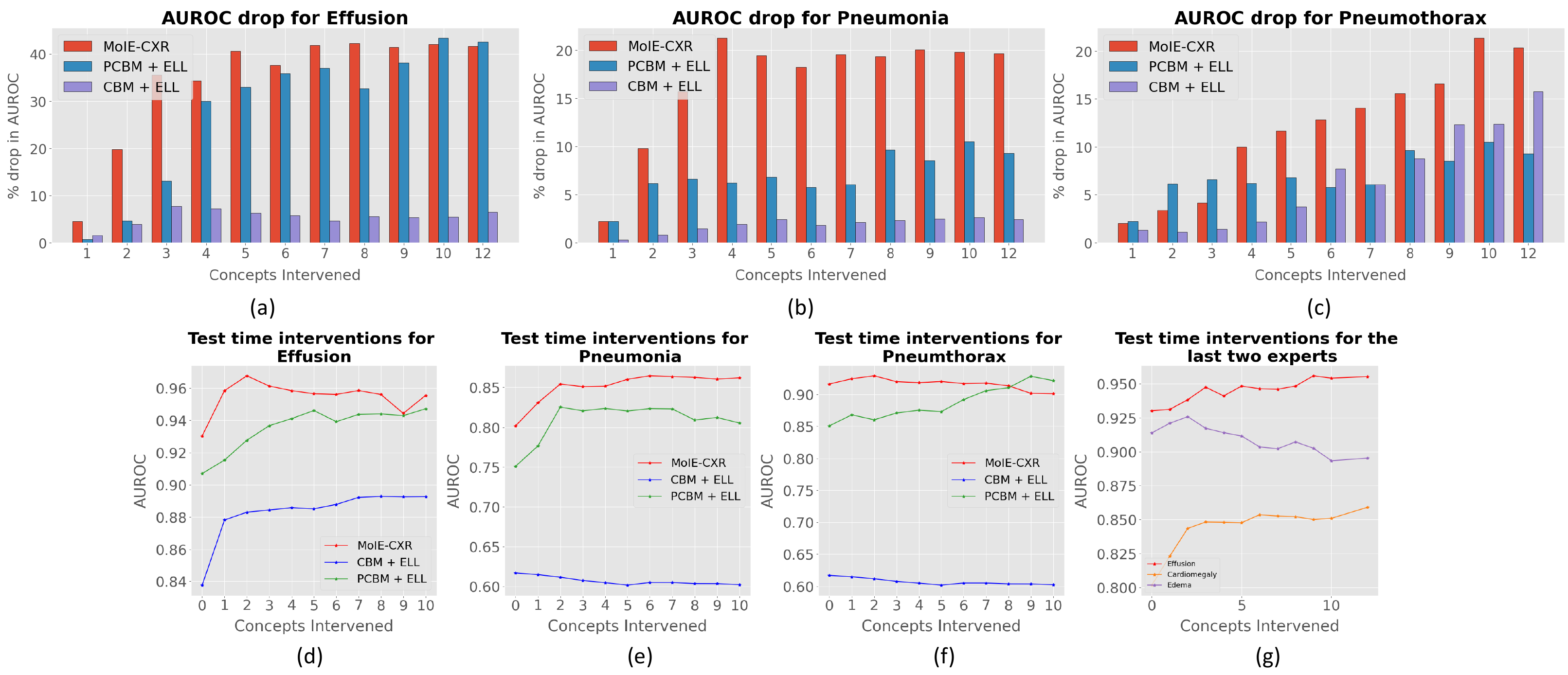}}
\caption{\textbf{(a-c):} Performance drop after zeroing out the concepts iteratively. The drop indicates the concepts to be more significant for prediction. \textbf{(d-g):} Test time interventions of concepts considering the ground truth concepts as an oracle on all samples (d-f), on the ``hard'' samples (g), covered by only the last two experts of MoIE-CXR.}
\label{fig:expert_performance_cv_vit}
\end{center}
\end{figure*}

\begin{figure*}[h]
\begin{center}
\centerline{\includegraphics[width=\linewidth]{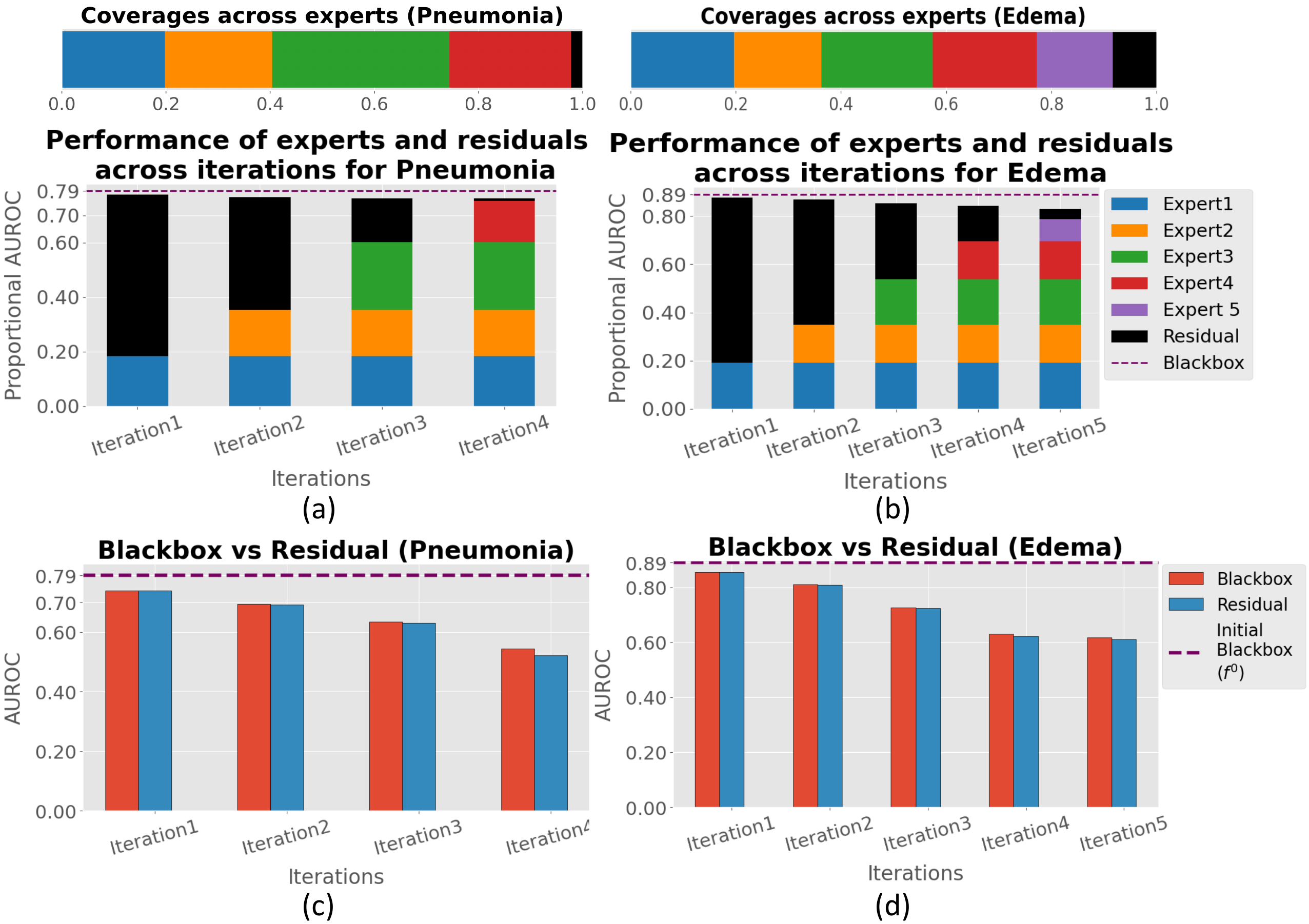}}
\caption{\textbf{(a-b)}: The performances of experts and residuals across iterations for pneumonia and edema. \textbf{(c-d)}: Performance comparison of the residuals and $f^0$ for the samples covered by the successive residuals.
}
\label{fig:expert_performance_cv_vit}
\end{center}
\end{figure*}

\begin{table}[H]
\caption{Hyperparameters of interpretable experts ($g$) for the dataset MIMIC-CXR.}
\label{tab:g_config_mimic_cxr}
\begin{center}
\begin{tabular}{l c c c c c }
\toprule 
    \thead{\textbf{Hyperparameter}} & 
    \thead{\textbf{Effusion}} & 
    \thead{\textbf{Cardiomegaly}} & 
    \thead{\textbf{Pneumothorax}} &
    \thead{\textbf{Pneumonia}} &
    \thead{\textbf{Edema}} \\
  
\midrule 
       Batch size & 1028 & 1028 & 1028 & 1028 & 1028   \\
       Learning rate & 0.01 & 0.01 & 0.01 & 0.01 & 0.01\\
       $\lambda_{lens}$ & 0.0001 & 0.0001 & 0.0001  & 0.0001 & 0.0001\\
       $\alpha_{KD}$ & 0.99 & 0.99 & 0.99 & 0.99 & 0.99 \\
       $T_{KD}$ & 20 & 20 & 20  & 20 & 20  \\
       hidden neurons & 30, 30 & 20, 20 & 20, 20 & 20, 20 & 20, 20 \\
       $\lambda_s$ & 96 & 1024 & 256 & 256 & 128   \\
       E-Lens ($T_{lens}$) & 7.6 & 7.6 & 10 & 10 & 7.6\\
       \# Expers ($T_{lens}$) & 5 & 4 & 5 & 4 & 5\\
\bottomrule
\end{tabular}
\end{center}
\end{table}

\label{sec:supplemental}
\end{document}